\pdfoutput=1

\documentclass[11pt]{article}

\usepackage[preprint]{acl}

\usepackage{times}
\usepackage{latexsym}

\usepackage[T1]{fontenc}

\usepackage[utf8]{inputenc}

\usepackage{microtype}

\usepackage{inconsolata}

\usepackage{graphicx}
\usepackage{amsmath,amsfonts}
\usepackage{algorithmic}
\usepackage{algorithm}
\usepackage{array}
\usepackage[caption=false,font=normalsize,labelfont=sf,textfont=sf]{subfig}
\usepackage{booktabs}
\usepackage{multirow}
\usepackage{textcomp}
\usepackage{stfloats}
\usepackage{url}
\usepackage{verbatim}
\usepackage{graphicx}
\usepackage{colortbl} 

\usepackage{tcolorbox} 

\usepackage{pifont}

\title{Multimodal Large Language Models for Text-rich Image Understanding: A Comprehensive Review}

\author{Pei Fu$^{1}$, Tongkun Guan$^{2}$, Zining Wang$^{1}$, Zhentao Guo$^{3}$, Chen Duan$^{1}$, \\
\textbf{Hao Sun$^{4}$, Boming Chen$^{1}$, Jiayao Ma$^{1}$, Qianyi Jiang$^{1}$, Kai Zhou$^{1}$, Junfeng Luo$^{1}$ } \\
  $^{1}$Meituan, $^{2}$ Shanghai Jiao Tong University, $^{3}$Beijing Institute of Technology, \\
  $^{4}$MAIS \& NLPR, Institute of Automation, Chinese Academy of Sciences\\
  \texttt{\{fupei,duanchen02,wangzining03,chenboming,majiayao02\}@meituan.com} \\
  \texttt{\{jiangqianyi02,zhoukai03,luojunfeng\}@meituan.com} \\
\texttt{gtk0615@sjtu.edu.cn,hao.sun@cripac.ia.ac.cn,zt\_guo1230@163.com}\
  }

\begin{document}
\twocolumn[{
\renewcommand\twocolumn[1][]{#1}
\maketitle
\vspace{-1em}
\begin{center}
    \captionsetup{type=figure}
    \includegraphics[width=0.98\textwidth]{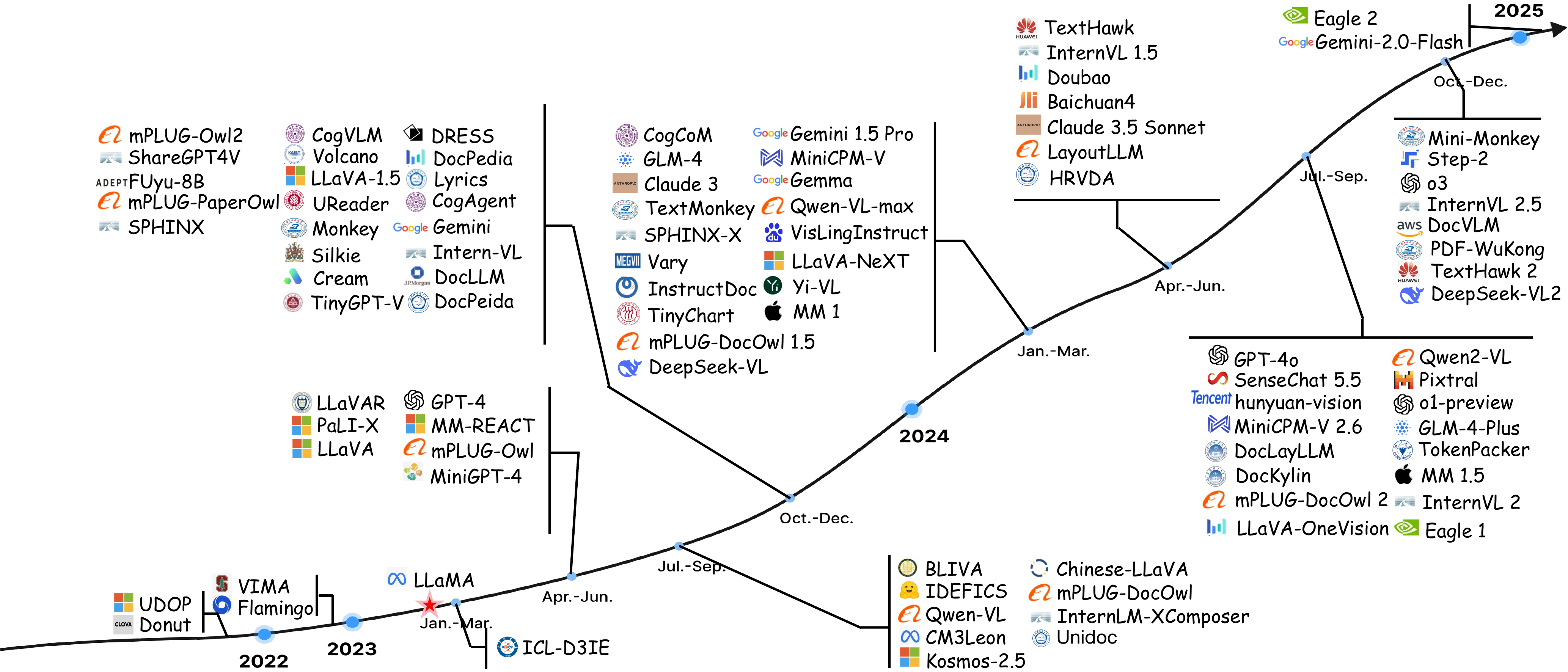}
    \captionof{figure}{The development timeline of TIU MLLMs. 
    }
    \label{fig:Development}
\end{center}
\vspace{1em}
}]

\begin{abstract}
The recent emergence of Multi-modal Large Language Models (MLLMs) has introduced a new dimension to the Text-rich Image Understanding (TIU) field, with models demonstrating impressive and inspiring performance. However, their rapid evolution and widespread adoption have made it increasingly challenging to keep up with the latest advancements. To address this, we present a systematic and comprehensive survey to facilitate further research on TIU MLLMs. Initially, we outline the timeline, architecture, and pipeline of nearly all TIU MLLMs. Then, we review the performance of selected models on mainstream benchmarks. Finally, we explore promising directions, challenges, and limitations within the field.

\end{abstract}
\section{Introduction}



Text-rich images play a pivotal role in real-world scenarios by efficiently conveying complex information and improving accessibility \cite{biten2019scene}. Accurately interpreting these images is essential for automating information extraction, advancing AI systems, and optimizing user interactions. To formalize this research domain, we term it \textbf{T}ext-rich \textbf{I}mage \textbf{U}nderstanding (\textbf{TIU}), which encompasses two core capabilities: perception and understanding. The perception dimension focuses on visual recognition tasks, such as text detection \cite{liao2022real}, text recognition~\cite{guan2025ccdplus}, formula recognition \cite{TRUONG2024110531, guan2024posformer}, and document layout analysis \cite{Yupan2022ARXIV_LayoutLMv3_Pre_training}. The understanding dimension, conversely, requires semantic reasoning for applications like key information extraction and document-based visual question answering (\emph{e.g.}, DocVQA \cite{mathew2021docvqa}, ChartQA \cite{masry2022chartqa}, and TextVQA~\cite{singh2019towards}).

Historically, perception and understanding tasks were handled separately through specialized models or multi-stage pipelines. Recent advances in vision-language models have unified these tasks within Visual Question Answering (VQA) paradigms, driving research towards the development of end-to-end universal models.

Figure \ref{fig:Development} presents an evolutionary timeline delineating critical milestones in unified text-rich image understanding models. The trajectory reveals two distinct eras: (a) The pre-LLM period (2019-2022) characterized by specialized architectures like LayoutLM \cite{Xu2019ARXIV_LayoutLM_Pre_training} and Donut \cite{Kim2021SEMANTIC_Donut_Document_Understanding}, which employed modality-specific pre-training objectives (masked language modeling, masked image modeling, \textit{etc.}) coupled with OCR-derived supervision (text recognition, spatial order recovery, \textit{etc.}). While effective in controlled settings, these models exhibited limited adaptability to open-domain scenarios due to their task-specific fine-tuning requirements and constrained cross-modal interaction mechanisms. (b) The post-LLM era (2023–present) is marked by the growing popularity of LLMs. Some studies propose Multimodal Large Language Models (MLLMs), which integrate LLM with visual encoders to jointly process visual tokens and linguistic elements through unified attention mechanisms, achieving end-to-end sequence modeling.

\begin{figure*}[t]
    \centering
    \includegraphics[width=\textwidth]{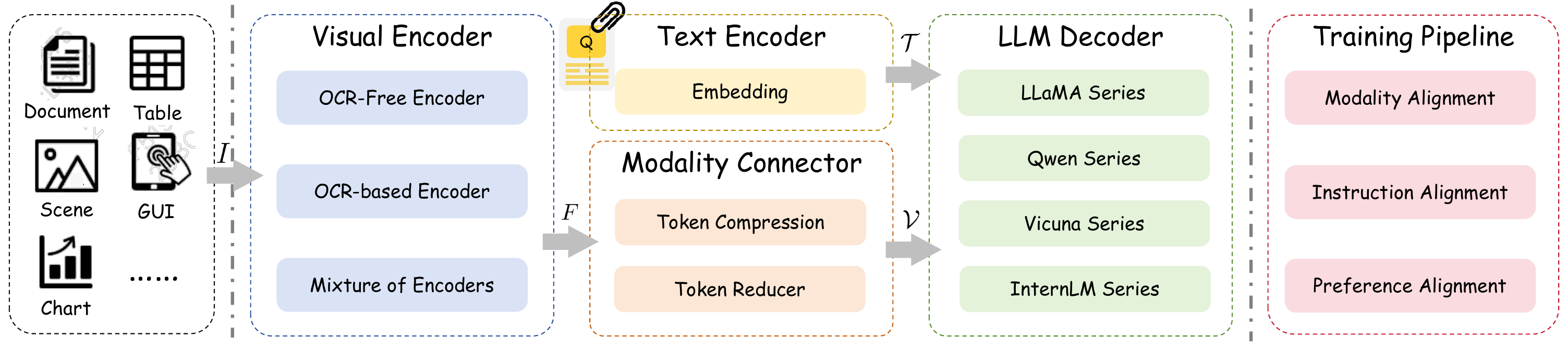} 
    \vspace{-2em}
    \caption{The general model architecture of MLLMs and the implementation choices for each component.} 
    \label{fig:Architecture} 
    \vspace{-1em}
\end{figure*}

This paradigm evolution addresses two critical limitations of earlier methods. First, the emergent MLLM framework eliminates modality-specific inductive biases through homogeneous token representation, enabling seamless multi-task integration. Second, the linguistic priors encoded in LLMs empower unprecedented zero-shot generalization and allow direct application to diverse tasks without task-specific tuning. 

Although these MLLMs present impressive and inspiring results, their rapid evolution and broad adoption have made tracking cutting-edge advancements increasingly challenging. Therefore, a systematic review that is tailored for documents to summarize and analyse these methods is in demand. However, 
existing surveys on text-rich image understanding often exhibit narrow focus: they either analyze domain-specific scenarios (e.g., tables and figures \cite{huang2024detection}, charts \cite{Huang2024ARXIV_From_Pixels_to, alshetairy2024transformersutilizationchartunderstanding}, forms \cite{abdallah2024transformers}) or emphasize unified deep learning frameworks \cite{subramani2011survey,ding2024deep}.

Our systematic survey addresses the gap by providing the first comprehensive analysis of nearly all TIU MLLMs in four dimensions: Model Architectures (Section \ref{sec:model_architecture}), Training Pipeline  (Section \ref{sec:training_pipeline}), Datasets and Benchmarks (Section \ref{sec:datasets}), Challenges and Trends (Section \ref{sec:summary}). This holds both academic and practical significance for advancing the field.

\begin{figure*}
    \centering
    \includegraphics[width=0.95\textwidth]{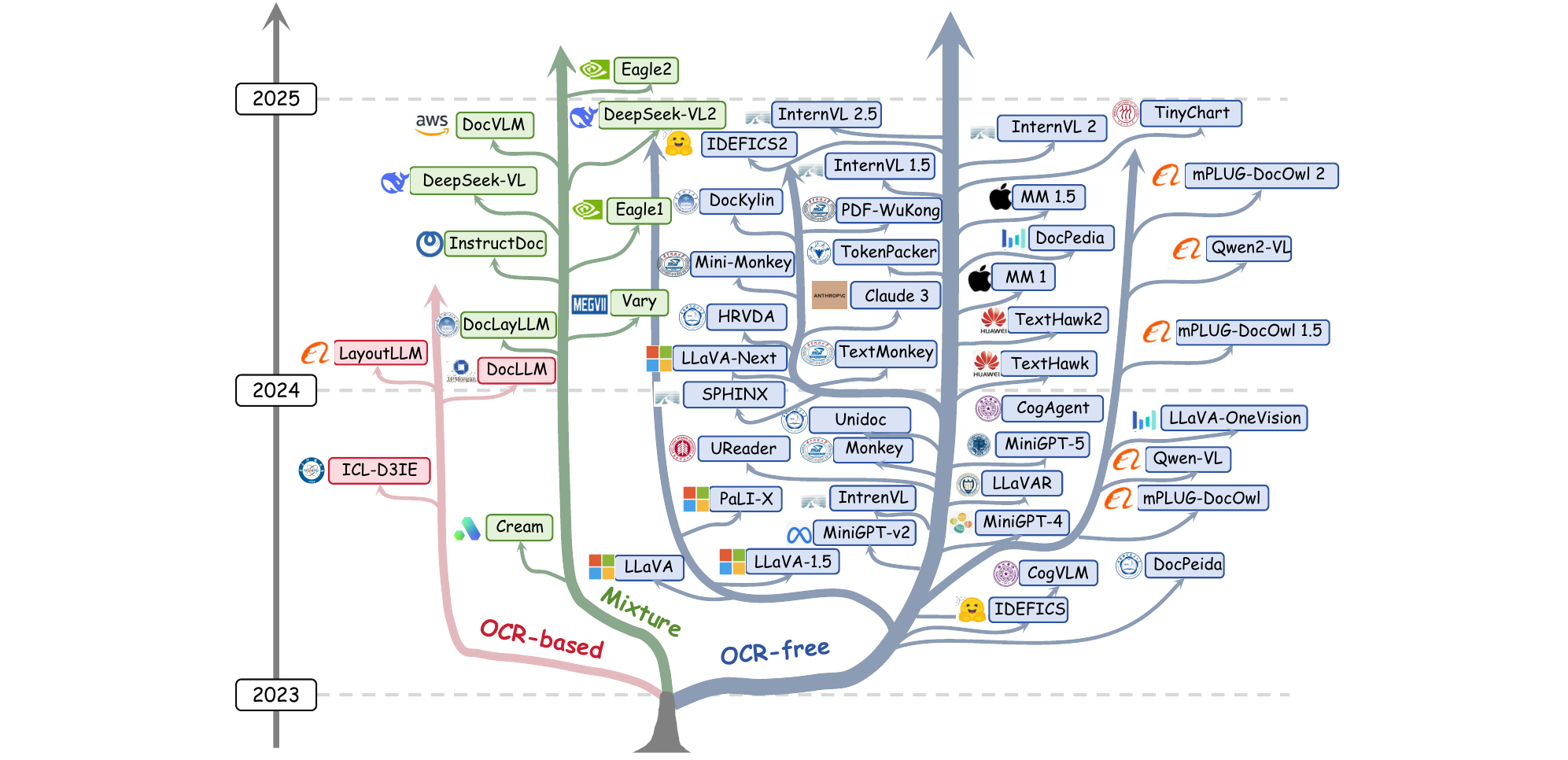}
    \vspace{-1em}
    \captionof{figure}{The evolutionary tree of modern LLMs traces the development of language models in recent years and highlights some of the most well-known models. According to the classification of Encoders, the \textcolor[HTML]{2E54A1}{blue} branch is ocr-free, the \textcolor[HTML]{C81D31}{pink} branch is ocr-based, and the \textcolor[HTML]{588F32}{green} branch is Mixture of Encoders.}
    \label{fig:2}
    \vspace{-1em}
\end{figure*}
\section{Model Architecture}
\label{sec:model_architecture}


TIU MLLM methods typically leverage pre-trained general visual foundation models to extract robust visual features or employ OCR engines to capture text and layout information from images. A modality connector is then used to align these visual features with the semantic space of the language features from the LLM. Finally, the combined visual-language features are fed into the LLM, which utilizes its powerful comprehension capabilities for semantic reasoning to generate the final answer.
As illustrated in Figure~\ref{fig:Architecture}, the framework of TIU MLLMs can be abstracted into three core components: Visual Encoder, Modality Connector, and LLM Decoder. 

\subsection{Visual Encoder}
\label{section_visual_encoder}
The Visual Encoder $\mathcal{F}(\cdot)$ is responsible for transforming input image $\mathbf{I}$ into feature representations $\mathbf{V}$, expressed as $\mathbf{V} = \mathcal{F}(\cdot)(\mathbf{I})$. As illustrated in Figure \ref{fig:2}, these encoders can be categorized into OCR-free, OCR-based, or a hybrid approach.

\noindent \textbf{OCR-free Encoder} is widely used to extract high-level visual features, effectively capturing essential information about objects, scenes, and textures. The commonly used OCR-free encoders include (1) \textbf{CLIP} \cite{radford2021CLIP}; (2) \textbf{ConvNeXt} \cite{woo2023convnext}; (3) \textbf{SAM} \cite{kirillov2023SAM}; (4) \textbf{DINOv2} \cite{oquab2023dinov2}; (5) \textbf{Swin-T} \cite{liu2021swin}; (6) \textbf{InternViT} \cite{chen2024internvlscalingvisionfoundation}.







\noindent \textbf{OCR-based Encoder} processes textual content and layout information from OCR outputs through three primary paradigms:
(1) \textbf{Direct Input} injects raw OCR texts into LLMs, though long sequences degrade inference efficiency \cite{he2023icl};
(2) \textbf{Cross-Attention} dynamically selects salient content via attention mechanisms within LLMs \cite{Wang2023ARXIV_DocLLM_A_layout};
(3) \textbf{External Encoder} employs specialized models like BLIP-2~\cite{li2023blip2bootstrappinglanguageimagepretraining}, DocFormerv2~\cite{nacson2024docvlmmakevlmefficient} or LayoutLMv3~\cite{Yupan2022ARXIV_LayoutLMv3_Pre_training} to structure OCR features before LLM integration \cite{ Tanaka2024AAAI_InstructDoc_A_Dataset, Luo2024CVPR_LayoutLLM_Layout_Instruction, Fujitake2024LREC_LayoutLLM_Large_Language}.

\noindent \textbf{Mixture of Encoders} strategies address TIU task complexity through two dominant configurations:
(1) \textbf{Dual OCR-Free} architectures (\emph{e.g.}, CLIP+SAM) combine complementary visual encoders to jointly capture global semantics and local details~\cite{wei2024vary};
(2) \textbf{Hybrid OCR-Free/OCR-Based} systems (\emph{e.g.}, CLIP+LayoutLMv3) synergize visual feature extraction with text-layout understanding, proving particularly effective for document-level tasks requiring multimodal reasoning \cite{Liao2024ARXIV_DocLayLLM_An_Efficient}.

\subsection{Modality Connector}
\label{section_connector}

Visual embeddings $\mathbf{V} = [\mathbf{v}_{1}, \mathbf{v}_{2},..., \mathbf{v}_{n}]$ and language embeddings $\mathbf{T} = [\mathbf{t}_{1}, \mathbf{t}_{2},..., \mathbf{t}_{l}]$ belong to different modalities. Consequently, to bridge the gap between them and create unified sequence representations that can be processed by large language models (LLMs), a modality connector $\xi: \mathbf{V} \xrightarrow{} \mathbf{T}$ is typically employed, which is responsible for converting $n$ visual features into $m$ visual tokens.
We review the strategies previously utilized in the literature for this purpose. 

Specifically, the modality connector can be easily implemented using a simple linear projector or multi-layer perception (MLP), \emph{i.e.,} $m=n$, but faces challenges in scalability and efficiency. Recent works also proposed more effective and innovative modality connectors from various perspectives, such as token compression and token reduction. The former focuses on reducing the number of inputs to the MLLM token with lossless compression, and the latter addresses the issue of costly tokens by removing redundant and unimportant token representations, such as background tokens.

\noindent \textbf{Token Compression}

1) Pixel shuffle \cite{chen2024far} rearranges the elements of a high-resolution feature map $(h, w)$ to form a lower-resolution feature map $(\frac{h}{s}, \frac{w}{s})$ by redistributing the spatial dimensions into the depth (channels) of the feature map. Here, $s$ denotes the compression rate. 
We summarized the process as $\xi: \mathbb{R}^{h \times w \times C} \xrightarrow{} \mathbb{R}^{\frac{h}{s} \times \frac{w}{s} \times (s\times C)}$.


\noindent \textbf{Token Reducer} 

1) Cross Attention \cite{alayrac2022flamingovisuallanguagemodel,li2023blip2bootstrappinglanguageimagepretraining,chen2024internvlscalingvisionfoundation,dai2023instructblipgeneralpurposevisionlanguagemodels} operates on the queries (a group of trainable vectors or the key features of the model itself) and the keys which are the image features produced by the vision encoder. We summarized the process as $\xi: \mathbb{R}^{h \times w \times C} \xrightarrow{} \mathbb{R}^{q \times D}$. 


2) H-Reducer~\cite{Hu2024ARXIV_mPLUG_DocOwl_1} introduces the $1 \times 4$ convolution layer to reduce visual features, as the horizontal texts are widely found in natural scenes and semantically coherent.
We summarized the process as $\xi: \mathbb{R}^{h \times w \times C} \xrightarrow{} \mathbb{R}^{h \times \frac{w}{4} \times D}$.

3) C/D-abstract~\cite{cha2024honeybee} employs Convolution and Deformable Attention respectively to achieve both flexibility and locality preservation.

4) Attention Pooling~\cite{Liu2024ARXIV_TextMonkey_An_OCR,Huang2024ARXIV_Mini_Monkey_Alleviating} identifies important tokens and removes redundant ones. To evaluate the redundancy of image features, the similarity between image tokens is often utilized~\cite{Liu2024ARXIV_TextMonkey_An_OCR}. This method selects tokens that are highly unique and lack closely similar counterparts. Average pooling is the most special one.
\begin{table*}[!t]
    \centering
    \scalebox{0.78}{
        \centering
    \scriptsize
    \begin{tabular}{l|>{\centering\arraybackslash}m{3.cm}>{\centering\arraybackslash}m{2.cm}>{\centering\arraybackslash}m{1.7cm}|>{\centering\arraybackslash}m{1.6cm}|>{\centering\arraybackslash}m{0.6cm}>{\centering\arraybackslash}m{0.6cm}>{\centering\arraybackslash}m{0.6cm}>{\centering\arraybackslash}m{0.8cm}|>{\centering\arraybackslash}m{0.6cm}}
    \toprule
    Model & Visual Encoder & Modality Connector & LLM Decoder & Training Pipline & DocVQA & InfoVQA & ChartQA & TextVQA & Avg. \\
   \midrule
   UReader~\cite{ye2023ureader} & CLIP-ViT-L/14 & Cross Attention & LLaMA-7B & MA+IA & 65.4 & 42.2 & 59.3 & 57.6 &56.13 \\
   DocLLM-1B~\cite{Wang2023ARXIV_DocLLM_A_layout} & - & - & Falcon-1B & MA+IA & 61.4 & - & - & - & -\\
    DocLLM-7B~\cite{Wang2023ARXIV_DocLLM_A_layout} & - & - & LLaMA2-7B & MA+IA & 69.5 & - & - & - &-\\

    Cream \cite{kim2023visually}& CLIP-ViT-L/14 & Cross Attention & Vicuna-7B & MA+IA & 79.5 & 43.5 & 63.0 & - &- \\
    
    LLaVA-13B~\cite{liu2023llava} & CLIP-ViT-L/14 & MLP & Vicuna-13B & MA+IA & 6.9 & - & - & 36.7 & -\\
    PaLI-X~\cite{chen2023pali} & ViT-22B & MLP & UL2-32B & MA+IA & 86.8 & 54.8 & 72.3 & 80.8  & 73.68\\
    LLaVAR~\cite{zhang2023llavar}& CLIP-ViT-L/14 & MLP & Vicuna-13B & MA+IA & 11.6 & - & - & 48.5 & -\\
    
    Qwen-VL~\cite{bai2023qwenvl} & ViT-bigG & Cross Attention & Qwen-7B & MA+IA & 65.1 & 35.4 & 65.7 & 63.8 &57.50\\
    LLaVA-1.5-7B~\cite{liu2023improvedllava} & CLIP-ViT-L & MLP & Vicuna1.5-7B & MA+IA & - & - & - & 58.2 &- \\
    LLaVA-1.5-13B~\cite{liu2023improvedllava} & CLIP-ViT-L & MLP & Vicuna1.5-13B & MA+IA & - & - & - & 62.5 &- \\
    CogAgent~\cite{hong2023cogagent} & EVA2-CLIP  & MLP+Cross Attention & Vicuna-13B & MA+IA & 81.6 & 44.5 & 68.4 & 76.1 & 67.65\\
    Unidoc~\cite{feng2023unidoc} & CLIP-ViT-L/14 & MLP & Vicuna-13B & MA+IA & 90.2 & 36.8 & 70.5 & 73.7 & 67.80 \\
    
    Monkey~\cite{Li2024CVPR_Monkey_Image_Resolution} & Vit-BigG & Cross Attention & Qwen-7B & MA+IA & 66.5 & 36.1 & 65.1 & 67.6 & 58.83\\
    Mini-Monkey~\cite{Huang2024ARXIV_Mini_Monkey_Alleviating} & InternViT-300M & MLP & InternLLM2-2B & IA & 87.4 & 60.1 & 76.5 & 75.7 & 74.93\\
    TextMonkey~\cite{Liu2024ARXIV_TextMonkey_An_OCR} & Vit-BigG & Cross Attention & Qwen-7B & MA+IA & 73.0 & - & 66.9 & 65.6 & -\\
    IDEFICS2 (\cite{laurençon2024mattersbuildingvisionlanguagemodels}) & SigLIP-SO400M & Cross Attention & Mistral-7B & MA+IA & 74.0 & - & - & 73.0 & -\\
    LayoutLLM~\cite{luo2024layoutllm} & LayoutLMv3-large & MLP & Vicuna1.5-7B & MA+IA & 74.25 & - & - & - & -\\
    DocKylin~\cite{zhang2024dockylin} & Swin & MLP & Qwen-7B & MA+IA & 77.3 & 46.6 & 66.8 & - & -\\
    DocLayLLM~\cite{liao2024doclayllm}  & LayoutLMV3 & MLP & LLaMA3-8B & MA+IA & 77.79 & 42.02 & - & - & -\\
    mPLUG-DocOwl ~\cite{Hu2024ARXIV_mPLUG_DocOwl_1}  & CLIP-ViT-L/14 & Cross Attention & LLaMA-7B & MA+IA & 62.2 & 38.2 & 57.4 & 52.6 & 52.60\\
    mPLUG-DocOwl1.5 ~\cite{hu2024mplug_docowl1.5_arxiv}  & CLIP-ViT-L/14  & H-Reducer & LLaMA2-7B & MA+IA & 82.2 & 50.7 & 70.2 & 68.6 & 67.93\\
    mPLUG-DocOwl2 ~\cite{Hu2024ARXIV_mPLUG_DocOwl2_High}  & CLIP-ViT-L/14  & H-Reducer & LLaMA2-7B & MA+IA & 80.7 & 46.4 & 70.0 & 66.7 & 65.95\\
    Vary~\cite{wei2024vary}  & CLIP-ViT-L/14 + SAM & MLP & Qwen-7B & MA+IA & 76.3 & - & 66.1 & - & -\\
    Eagle~\cite{shi2024eagleexploringdesignspace}  & CLIP + ConvNeX + Pix2Struct + EVA2 + SAM & MLP & LLaMA3-8B & MA+IA & 86.6 & - & 80.1 & 77.1 & -\\
    
    PDF-WuKong~\cite{Xie2024ARXIV_PDF_WuKong_A} & CLIP-ViT-L-14 & Cross Attention & InernLM2-7B & MA+IA & 85.1 & 61.3 & 80.0 & - & -\\
    InstructDoc~\cite{tanaka2024instructdoc} & CLIP/Eva-CLIP-ViT & Cross Attention + MLP & Flan-T5/OPT & MA+IA & - & 50.9 & 29.4 & 53.8 &-\\
    
    TextHawk~\cite{yu2024texthawk}  & SigLIP & Cross Attention & InternLM-XComposer & MA+IA & 76.4 & 50.6 & 66.6 & -&-\\
    TextHawk2~\cite{yu2024texthawk2}  & SigLIP & Cross Attention & Qwen2-7B & MA+IA & 89.6 & 67.8 & 81.4 & 75.1 &78.48\\
    MM1.5-1B~\cite{zhang2024mm1_5} & CLIP-ViT-H & C-Abstractor & Private & MA+IA & 81.0 & 50.5 & 67.2 & 72.5 &67.80\\
    MM1.5-3B~\cite{zhang2024mm1_5} & CLIP-ViT-H & C-Abstractor & Private & MA+IA & 87.7 & 58.5 & 74.2 & 76.5 & 74.23\\
    MM1.5-7B~\cite{zhang2024mm1_5} & CLIP-ViT-H & C-Abstractor & Private & MA+IA & 88.1 & 59.5 & 78.6 & 76.5 & 75.68\\
    MM1.5-30B~\cite{zhang2024mm1_5} & CLIP-ViT-H & C-Abstractor & Private & MA+IA & 91.4 & 67.3 & 83.6 & 79.2 & 80.38\\
    HRVDA~\cite{liu2024hrvda} & Swin-L & MLP & LLaMA2-7B & MA+IA & 72.1 & 43.5 & 67.6 & 73.3 & 64.13\\
    InternVL1.5-26B~\cite{chen2024far} & InternViT-6B & Pixel-shuffle + MLP & InternLM2-20B & MA+IA & 90.9 & 72.5 & 83.8 & 80.6 & 81.95\\
    InternVL2.5-1B~\cite{chen2024internvl2_5} & InternViT-300M & Pixel-shuffle + MLP & Qwen2.5-0.5B & MA+IA & 84.8 & 56.0 & 75.9 & 72.0 & 72.18\\
    InternVL2.5-2B~\cite{chen2024internvl2_5} & InternViT-300M & Pixel-shuffle + MLP & InternLM2.5-1.8B & MA+IA & 88.7 & 60.9 & 79.2 & 74.3 & 75.78\\
    InternVL2.5-4B~\cite{chen2024internvl2_5} & InternViT-300M & Pixel-shuffle + MLP & Qwen2.5-3B & MA+IA & 91.6 & 72.1 & 84.0 & 76.8 & 81.13 \\
    InternVL2.5-8B~\cite{chen2024internvl2_5} & InternViT-300M & Pixel-shuffle + MLP & InternLM2.5-7B & MA+IA & 93.0 & 77.6 & 84.8 & 79.1 & 83.63 \\
    InternVL2.5-26B~\cite{chen2024internvl2_5} & InternViT-6B & Pixel-shuffle + MLP & InternLM2.5-20B & MA+IA & 94.0 & 79.8 & 87.2 & 82.4 & 85.85 \\
    InternVL2.5-38B~\cite{chen2024internvl2_5} & InternViT-6B & Pixel-shuffle + MLP & Qwen2.5-32B & MA+IA & 95.3 & 83.6 & 88.2 & 82.7 & 87.45\\
    InternVL2.5-78B~\cite{chen2024internvl2_5} & InternViT-6B & Pixel-shuffle + MLP & Qwen2.5-72B & MA+IA & 95.1 & 84.1 & \cellcolor{gray!30}\textbf{88.3} & 83.4 & 87.73 \\
    InternVL2.5-8B-mpo\cite{wang2024enhancingreasoningabilitymultimodal}$\dagger$  & InternViT-300M & Pixel-shuffle +MLP & InternLM2.5-7B & PA & 92.3 & 76.0 & 83.8 & 79.1 & 82.80 \\
    DocPeida~\cite{Feng2024ARXIV_DocPedia_Unleashing_the} & Swin & MLP & Vicuna-7B & MA+IA & 47.1 & 15.2 & 46.9 & 60.2 & 42.35 \\
    TinyChart~\cite{zhang2024tinychart} & SigLIP & MLP & Phi-2 & IA & - & - & 83.6 & - & - \\
    TokenPacker-7B~\cite{li2024tokenpackerefficientvisualprojector} & CLIP-ViT-L/14 & Cross Attention & Vicuna-7B & MA+IA & 60.2 & - & - & - & - \\
    TokenPacker-13B~\cite{li2024tokenpackerefficientvisualprojector} & CLIP-ViT-G/14 & Cross Attention & Vicuna-13B & MA+IA & 70.0 & - & - & - & - \\
    
    LLaVA-OneVision-0.5B~\cite{li2024llava} & SigLIP & MLP & qwen2-0.5B & MA+IA & 70.0 & 41.8 & 61.4 & - & - \\
    LLaVA-OneVision-7B~\cite{li2024llava} & SigLIP & MLP & qwen2-7B & MA+IA & 87.5 & 68.8 & 80.0 & - & - \\
    Qwen2-VL-2B~\cite{wang2024qwen2vl} & CLIP-VIT-G/14 & Cross Attention & Qwen2-2B & MA+IA & 90.1 & 65.5 & 73.5 & 79.7 & 77.20 \\ 
    Qwen2-VL-7B~\cite{wang2024qwen2vl} & CLIP-VIT-G/14 & Cross Attention & Qwen2-7B & MA+IA & 94.5 & 76.5 & 83.0 & 84.3 & 84.58 \\ 
    DocVLM~\cite{nacson2024docvlmmakevlmefficient} & CLIP-ViT-G/14 + DocFormerV2 & Cross Attention & Qwen2-7B & MA+IA & 92.8 & 66.8 & - & 82.8 & - \\
    Qwen2-VL-72B~\cite{wang2024qwen2vl} & CLIP-VIT-G/14 & Cross Attention & Qwen2-72B & MA+IA & \cellcolor{gray!30}\textbf{96.5} & \cellcolor{gray!30}\textbf{84.5} & \cellcolor{gray!30}\textbf{88.3} & \cellcolor{gray!30}\textbf{85.5} & \cellcolor{gray!30}\textbf{88.70} \\ 

    DeepSeek-VL2-3B~\cite{wu2024deepseek} & SigLIP-SO400M-384 & Pixel-shuffle + MLP & DeepSeekMoE & MA+IA &88.9 &66.1 &81.0 &80.7 &79.18 \\
    DeepSeek-VL2-16B~\cite{wu2024deepseek} & SigLIP-SO400M-384 & Pixel-shuffle + MLP & DeepSeekMoE & MA+IA &92.3 &75.8 &84.5 &83.4 &84.00 \\
    DeepSeek-VL2-27B~\cite{wu2024deepseek} & SigLIP-SO400M-384 & Pixel-shuffle + MLP & DeepSeekMoE & MA+IA &93.3 &78.1 &86.0 &84.2 &85.40 \\
    
    Eagle2~\cite{li2025eagle2buildingposttraining} & SigLIP + ConvNeXt & MLP & Qwen2.5-7B & MA+IA & 92.6 & 77.2 & 86.4 & 83.0 & 84.80 \\
    \bottomrule
    \end{tabular}


    }
    \vspace{-0.5em}
    \caption{The summary of representative mainstream MLLMs, including the model architectures, training pipelines, and scores on the four most popular benchmarks of TIU. ``Private'' indicates that the MLLM utilizes a proprietary large model. ``$\dagger$'' indicates the results are obtained by downloading official open-source model and testing it locally.}
    \label{tab:mllm_summary}
    \vspace{-1.5em}
\end{table*}

\subsection{LLM Decoder}
The aligned features are fed into the LLM decoder together with the language embeddings for reasoning. We list the commonly used LLMs in MLLM:

\noindent \textbf{LLaMA Series}. LLaMA \cite{touvron2023llama, touvron2023llama2, dubey2024llama} is a series of open-source large language models developed by Meta, aimed at promoting openness and innovation in artificial intelligence technology, LLaMA series include models of varying parameter scales (e.g., 7B, 13B, 34B). 

\noindent \textbf{Qwen Series}. Qwen \cite{bai2023qwen,yang2024qwen2}, developed by Alibaba, is a multilingual LLM that supports both Chinese and English.

\noindent \textbf{Vicuna Series}. Vicuna \cite{zheng2023judging} is an open-source large language model built on LLaMA, developed by research teams from institutions including UC Berkeley, CMU, and Stanford. 

\noindent \textbf{InternLM series}. InternLM \cite{cai2024internlm2} is an open-source large language model series developed by the Shanghai Artificial Intelligence Laboratory, with the latest version, InternLM 2.5, offering parameter sizes of 1.8B, 7B, and 20B. 
\section{Training Pipeline}
\label{sec:training_pipeline}
The training pipeline of MLLM for TIU can be delineated into three main stages: 1) Modality Alignment (\textbf{MA}); 2) Instruction Alignment (\textbf{IA}); and 3) Preference Alignment ({\textbf{PA}}).

\subsection{Modality Alignment}
In this stage, previous works typically use OCR data from traditional OCR tasks~\cite{guan2024bridging,guan2023ccd} to pre-train the MLLM, which aims to bridge the modality gap. The general alignment methods can be categorized into three types: recognition, localization, and parsing.

\noindent \textbf{Read Full Text.} UReader~\cite{ye2023ureader} is the first to explore unified document-level understanding, which introduces the Read Full Text task in VQA for pre-training. Specifically, they include 1) reading all texts from top to bottom and left to right, and 2) reading the remaining texts based on given texts.
Compared to reading the full text, some works~\cite{Lv2024ARXIV_KOSMOS_2_5, Hu2024ARXIV_mPLUG_DocOwl_1} proposes a more structured reading approach by predicting the image markdown, not text transcriptions.

\noindent \textbf{Reading Partial Text within Localization.} Due to the length of document texts, instructions for reading the full text may risk truncation because of the limited token length in LLMs. To address these limitations, Park \emph{et al.}\cite{park2024hierarchical} introduced two novel tasks: Reading Partial Text (RPT) and Predicting Text Position (PTP). The former randomly selects and reads continuous portions of text in the reading order from top to bottom and left to right. For example, ``Q: What is the text in the image between the first 30\%, from 20\% to 40\%, or the last 16\%?'' For the PTP task, given a text segment, the MLLM aims to infer its relative position (percentage format) within the full text. 
For example, ``Q: Where is the text {query texts} located within the image? A: 40\% to 80\%''.
However, this approach can be somewhat obscure and challenging to express accurately. 

Alternatively, some methods~\cite{hu2024mplug_docowl1.5_arxiv,yu2024texthawk,liu2024hrvda} extract texts based on specific spatial positions, which are summarized into two types.
1) Text Recognition aims to extract the textual content from a given position in the image, ensuring that the model can accurately recognize and extract text within specific regions.
2) Text Grounding involves identifying the corresponding bounding box for specific text in the image, which assists the model in understanding the document layout.



\noindent \textbf{Parsing.} In document images, many elements (charts, formulas, and tables) may not be represented using plain text. An increasing number of researchers are now focusing on these element parsing. 
1) Chart Parsing. Chart types include vertical bars, horizontal bars, lines, scatter plots, and pie charts. Charts serve as visual representations of tables, and organizing text in reading order fails to capture their structure. To preserve their mathematical properties, researchers often convert charts into tables. This process involves breaking down the chart into x/y axes and their corresponding values, which can be represented in Markdown, CSV formats, or even converted into Python code. This approach enables models to better understand the chart's specific meaning.

2) Table Parsing. Compared to charts, tables have a more standardized structure, where rows and columns form key-value pairs. Common formats for representing tables include LaTeX, Markdown, and HTML. Markdown is often used for simple tables due to its concise text format, while HTML can handle cells that span multiple rows and columns, despite its use of many paired tags like <tr></tr> and <td></td>. Some tables, with complex spanning, custom lines, spacing, or multi-page length, require LaTeX for representation. However, the diversity in LaTeX representations can make these tables challenging for models to fully understand.

3) Formula Parsing. Besides tables and charts, formulas are also commonly used. In the pre-training phase, models learn the LaTeX representation of formula images, enhancing their understanding of formulas. This provides a solid foundation for tasks involving formula computation and reasoning during the instruction alignment.

\subsection{Instruction Alignment}

Upon completing the modality alignment pre-training stage, the MLLM acquires basic visual recognition and dialogue capabilities. However, to achieve human-aligned intelligence, three critical capability gaps must be addressed:(1) Advanced multimodal perception and cross-modal reasoning abilities; (2) Prompt robustness across diverse formulations; (3) Zero-shot generalization for unseen task scenarios. To bridge these gaps, instruction alignment through supervised fine-tuning (SFT) has emerged as an effective paradigm. This phase typically unfreezes all model parameters and employs instructional data with structured templates. 

To systematically address these challenges, we have categorized the current methods emerging in instruction alignment into three distinct levels:

\noindent \textbf{1) Level 1: Visual-Semantic Anchoring.}  We categorize these instructions into two types: i) Answer within the image; and ii) Answer without the image.
This type of instruction data where answers are located directly within the image, assists MLLMs refine their accuracy in generating responses that are directly linked to specific visual content, reducing reliance on generic or contextually weak answers~\cite{mathew2021docvqa, mathew2022infographicvqa}.
Certain tasks require reasoning based on world knowledge and involve complex inference procedures, such as scientific question answering~\cite{masry2022chartqa,chen2021geoqa}. Consequently, these instructions are designed with the common characteristic that the answer is not directly visible in the image. This encourages the model to utilize its linguistic comprehension and external knowledge, enhancing its advanced reasoning and inference capabilities. An example might be: ``Q: How much higher is the red bar compared to the yellow bar in the chart, in terms of percentage? A: 12.1\%.''

\noindent \textbf{2) Level 2: Prompt Diversity Augmentation.} To bolster robustness in handling a broader spectrum of prompts, rather than being limited to specific prompts tailored for particular tasks, researchers often employ data augmentation on the question component of the instruction stream. A popular strategy involves leveraging existing large language models to rephrase the same question in multiple ways. For example, consider the original question: ``What is written on the sign in the image?'' It can be rephrased as: ``Can you read the text displayed on the sign shown in the image?''``Identify the sign in the image.''``Please examine the image and list the words that appear within the sign.''
By utilizing such varied templates, researchers can train MLLMs to better interpret and respond to a wide range of prompts, thereby enhancing their flexibility and accuracy in real-world applications.

\noindent \textbf{3) Level 3: Zero-shot Generalization.} To enhance the generalization ability to handle unseen tasks, several strategies typically are employed:

Chain of Thought (CoT)~\cite{wei2022chain} reasoning involves breaking down complex problems into a series of intermediate steps or sub-tasks, allowing a model to tackle each part systematically. Some studies have demonstrated improvements by incorporating text-level CoT reasoning~\cite{zhang2024cfret} or box-level visual CoT supervision~\cite{shao2025visual}. To better illustrate the process, consider the prompt: ``What is the average of the last four countries' data?'', the CoT reasoning unfolds as follows: i) Identify the data for the last four countries;
ii) Calculate the sum of these values;
iii) Calculate the average by dividing the sum by the number of countries.

Another strategy is Retrieval-Augmented Generation (RAG). RAG~\cite{arslan2024survey} combines the strengths of retrieval-based and generation-based approaches by integrating an information retrieval component with a generative model. This method allows the model to access a vast external knowledge base, retrieving pertinent information to inform and enhance the generation process. 

\begin{table*}[!t]
    \centering
    \scalebox{0.9}{
    \scriptsize
    \begin{tabular}{l|>{\centering\arraybackslash}m{4.2cm}|>{\centering\arraybackslash}m{1.2cm}|>{\centering\arraybackslash}m{3.5cm}|>{\centering\arraybackslash}m{1cm}|>{\centering\arraybackslash}m{1cm}|>{\centering\arraybackslash}m{1.5cm}}
    \toprule
    Domain & Dataset & Language & Scene Sources & \#Images & \#Q\&A pairs & Train/Test \\
    \midrule
    \multirow{25}{*}{Document} & DocVQA~\cite{mathew2021docvqa} & English & Industry document & 12,767&  50,000 & Train + Test \\
    & Docmatix~\cite{laurenccon2024building} & English & Industry document & 2.4M & 9.5M & Train \\
    & InfoVQA~\cite{mathew2022infographicvqa} & English & Infographics & 5,485 & 30,035 & Train + Test \\
    & MP-DocVQA~\cite{tito2023hierarchical} & English & Industry documents & 47,952 & 46,176 & Train + Test \\
    &DocGenome~\cite{xia2024docgenome} & English & Scientific document & 6.8M & 3,000 & Train \\
    &IIT-CDIP~\cite{xu2020layoutlm} & English & Multi-domain & 11M & - & Train \\
    &synthdog~\cite{kim2022ocr}& English & Multi-domain & 2M & - & Train \\
    &CCPdf~\cite{turski2023ccpdf}& Multilingual & Multi-domain & 1.1M & - & Train \\
    &RVL-CDIP~\cite{harley2015evaluation} & English & Industry document & 159,418 & - & Train \\
    &VisualMRC~\cite{tanaka2021visualmrc} & English & Webpage Document & 10,197 & 30,562 & Train + Test \\
    &KLC~\cite{stanislawek2021kleister} & English & Industry document & 2463 & 22,224 & Train + Test \\
    & OCREval~\cite{lv2023kosmos} & English & Multi-domain & 2,297 & - & Test \\
    & MMLongBench-Doc~\cite{Ma2024NEURIPS_MMLongBench_Doc_Benchmarking} & English & Multi-domain Long Documents & 135 & 1k & Test \\
    & Do-GOOD~\cite{He2023ARXIV_Do_GOOD_Towards} & English & Industry document & 410k & 50k & Test\\
    & OCR-VQA~\cite{mishra2019ocr} & English & Book covers & 207,572 & >1M & Train + Test\\
    & SlideVQA~\cite{tanaka2023slidevqa} & English & Slide decks & 52,480 & 14,484 & Train + Test\\
    & PDF-VQA~\cite{10.1007/978-3-031-43427-3_35} & English & Scientific document  & 13,484 & 140,610 & Train + Test\\
    & BenthamQA~\cite{mathew2021asking} & English & Handwritten document  & 338 & 200 & Train + Test\\
    & FinanceQA~\cite{Sujet-Finance-QA-Vision-100k} & English & Financial reports & 9,801 & 100k & - \\
    & Ureader~\cite{ye-etal-2023-ureader} & English & Multi-domain & 24.5k & 24.5k & Train \\
    & ColPali~\cite{faysse2024colpali} & English & Multi-domain & 118,695 & 118,695 & Train + Test \\
    & FUNSD~\cite{jaume2019funsd} & English & Scanned forms & 199 & 5312 & Train + Test \\
    & SROIE~\cite{huang2019icdar2019sroie} & English & Multi-domain & 973 & 52,316 & Train + Test \\
    & POIE~\cite{kuang2023visual} & English & Multi-domain & 3,000 & 111,155 & Train + Test \\
    & IAM~\cite{marti2002iam} & English & Lancaster-Oslo/Bergen & 1066 & - & Train \\
    \midrule 
    \multirow{12}{*}{Chart}& ChartQA~\cite{masry2022chartqa} & English & Charts and Plots  & 20,882 & 32,719 & Train + Test\\
    & PlotQA~\cite{methani2020plotqa} & English & Plots (Real world data source) & 224,377 & 28.9M & Train \\
    & FigureQA~\cite{kahou2017figureqa} & English & Science style image & >100,000 & >1.3M& Train \\
    & DVQA~\cite{kafle2018dvqa} & English & Data Visualizations & 300,000 & 3,487,194 & Train \\
    & Unichart~\cite{masry2023unichart} & English & Multi-domain & 290,736 & 300,000 & Train \\
    & LRV-Instruction~\cite{liu2023mitigating} & English & Multi-domain & 400k & 400k & Train + Test  \\
    & VisText~\cite{tang-etal-2023-vistext} & English & Financial reports & 12,441  & 12,441 & Train + Test \\
    & Chart2Text~\cite{obeid-hoque-2020-chart} & English & Financial reports & 8,305 & 8,305 & Train + Test \\
    & ArxivQA~\cite{li-etal-2024-multimodal-arxiv} & English & Scientific Chart & 35,000 & 100,000 & Train + Test \\

    & ChartY~\cite{chen2024onechart} & Multilingual & Charts and Plots & 6k & 6k & Test \\
    & ChartX~\cite{xia2024chartx} & English & Charts and Plots & 6k & 6k & Test\\
    & MMC~\cite{Liu2024NAACL_MMC_Advancing_Multimodal} & English & Plots (Real world data source) & 1.7k& 2.9k & Test\\
    & ChartBench~\cite{Xu2024ARXIV_ChartBench_A_Benchmark} & English & Plots (Real world data source) & 68k& 549k & Test \\
    
    \midrule 
    \multirow{13}{*}{Scene} & TextCaps~\cite{sidorov2020textcaps} & English & Scene Text & 28,408 & 142,040 & Train + Test \\
    & TextVQA~\cite{singh2019towards} & English & Scene Text & 28,408 & 45,336 & Train + Test \\
    & ST-VQA~\cite{biten2019scene} & English & Scene Text & 23,038 & 31,791 & Train + Test \\
    & MT-VQA~\cite{wen2024mt} & Multilingual & 20 fine-grained scenes & 8,794 & 28,607 & Train + Test \\
    & OCRVQA~\cite{mishra2019ocr} & English & Scene Text & 207,572 & 1M & Train \\
    & ICDAR13~\cite{karatzas2013icdar} & English & Scene Text & 229 & - & Train \\
    & ICDAR15~\cite{karatzas2015icdar} & English & Scene Text & 1000 & - & Train \\
    & TotalText~\cite{ch2017totaltext} & English & Scene Text & 1000 & - & Train \\
    & CTW1500~\cite{yuliang2017detecting} & English & Scene Text & 1255 & - & Train \\
    & LSVT~\cite{sun2019icdar} & Chinese & Scene Text & 30,000 & - & Train \\
    & RCTW~\cite{shi2017icdar2017} & Chinese & Scene Text & 8,034 & - & Train \\
    &LAION-OCR~\cite{schuhmann2022laion} & English & Scene Text & - & - & Train \\
    &Wukong-OCR~\cite{gu2022wukong} &Chinese & Scene Text & - & - & Train\\
    \midrule 
    \multirow{13}{*}{Table} & TableQA~\cite{sun2020tableqa} & English & Financial reports & 6,000 & 64,891 & Train \\
    &WikiTableQuestions~\cite{pasupat2015compositional}& English & Multi-domain & 2,108 & 22,033 & Train + Test\\
    &DeepForm~\cite{svetlichnaya2020deepform}& English & Political campaign finance receipts & 1100 & 5500 & Train + Test \\
    &TabFact~\cite{chen2019tabfact} & English & Wikipedia tables & 14,922 & 117,273 & Train + Test \\
    &TabMWP~\cite{lu2022dynamic} & English & Educational documents & 38,431 & 38,431 & Train \\
    &TURL~\cite{deng2022turl} & English & Wikipedia & 200,000 & - & Train \\
    &PubTabNet~\cite{zhong2020image} & English & Scientific articles & 200,000 & - & Train \\
    & TableVQA-Bench~\cite{Kim2024ARXIV_TableVQA_Bench_A} & English & Scientific and Financial Reports & 0.9k & 1.5k & Test \\
    & MMTab-eval~\cite{zheng2024multimodal} & English & Scientific and Financial Reports & 23k & 49k & Test \\
    & ComTQA~\cite{Zhao2024NEURIPS_TabPedia_Towards_Comprehensive} & English & Scientific and Financial Reports & 1.6k & 9k & Test \\
    & TAT-DQA~\cite{zhu2022towards} & English & Financial reports & 3,067 & 16,558 & Train + Test \\
    & VQAonBD~\cite{raja2023icdar} & English & Financial reports & 48,895 & 1,531,455 & Train + Test \\
    & MultiHiertt~\cite{zhao-etal-2022-multihiertt} & English & Financial reports & 89,646 & 10,440 & Train + Test \\

    \midrule 
    \multirow{2}{*}{GUI} & ScreenQA~\cite{hsiao2022screenqa} & English & Mobile app screenshots & 35,352 & 85,984 & Train + Test\\
    & Screen2Words~\cite{wang2021screen2words} & English & Android app screenshot & 22,417 & 112,085 & Train + Test \\
    \midrule 
    \multirow{8}{*}{Comprehensive} & OCRbench~\cite{liu2024ocrbench} & English & Multi-domain & 0.9k & 1k & Test\\
    & Seed-bench-2-plus~\cite{li2024seed} & English & Multi-domain & 0.6k & 2.3k & Test \\
    & CONTEXTUAL~\cite{wadhawan2024contextual} & English & Multi-domain & 0.5k & 0.5k & Test \\
    & OCRBench v2~\cite{fu2024ocrbench} & English & Multi-domain & 9.5k & 10k & Test\\
    & FOX~\cite{liu2024focus} & Multilingual & Scientific document  & 0.7k & 2.2k & Test\\
    & DocLocal4K~\cite{hu-etal-2024-mplug} & English & Multi-domain & 4.2k & 4.2k & Test\\
    & CC-OCR~\cite{yang2024cc} & Multilingual & Multi-domain & 7k & - & Test\\
    & MMDocBench~\cite{Zhu2024ARXIV_MMDocBench_Benchmarking_Large} & English & Multi-domain & 2.4k & 4.3k & Test\\
    \bottomrule
    \end{tabular}

    }
    \vspace{-0.5em}
    \caption{Representative datasets and benchmarks for Text-rich Image Understanding. Each dataset is marked for training and testing typically according to its content, functions, and user requirements. 
    }
    \label{tab:datawithbenchmark}
    \vspace{-1.5em}
\end{table*}

\subsection{Preference Alignment}
In the modality and instruction alignment stages, the model predicts the next token based on previous ground-truth tokens during training, and on its own prior outputs during inference. If errors occur in the outputs, this can lead to a distribution shift in inference. The more output the model has, the more serious this phenomenon becomes. In previous natural language processing (NLP) works~\cite{lai2024step,pang2025iterative}, a series of preference alignment techniques~\cite{rafailov2024direct, ouyang2022training,shao2024deepseekmath,wang2024mdpo} have been proposed to optimize the output of the model to make it more consistent with human values and expectations. Benefiting from the success of preference alignment applied to NLP, InternVL2-MPO~\cite{wang2024enhancing} introduces preference alignment to the multimodal field and proposes a Mixed Preference Optimization (MPO) to improve multimodal reasoning. Specifically, they propose a continuation-based Dropout Next Token Prediction (DropoutNTP) pipeline for samples lacking clear ground truth and a correctness-based pipeline for samples with clear ground truth. This strategy improves the performance of the model on OCRBench~\cite{fu2024ocrbench}. Nevertheless, its potential to enhance document multimodal reasoning remains under-explored.

\section{Datasets and Benchmarks}
\label{sec:datasets}


The rapid advancements in TIU tasks have been fundamentally driven by the proliferation of specialized datasets and standardized benchmarks. As illustrated in Table \ref{tab:datawithbenchmark}, we systematically categorize TIU-related datasets into two types: \textit{domain-specific} (Document, Chart, Scene, Table, and GUI) and \textit{comprehensive} scenarios. 


Specifically, some datasets are derived by converting training data from traditional tasks~\cite{guan2022industrial,guan2023self} into Visual Question Answering (VQA) formats, such as text detection, text spotting, table recognition, and \emph{etc.}. These datasets are typically utilized for modality alignment in the first stage of training, enabling models to bridge the gap between textual and visual information effectively. Other datasets are specifically designed in VQA formats for certain scenarios, such as DocVQA~\cite{mathew2021docvqa}, InfoVQA~\cite{mathew2022infographicvqa}, ChartQA~\cite{masry2022chartqa}, and TextVQA~\cite{singh2019towards}. These datasets have played a pivotal role in advancing the field of TIU by providing structured and domain-specific challenges. Their introduction has significantly accelerated progress in tasks like document understanding, chart interpretation, and natural scene text comprehension. Consequently, published papers frequently report these metrics, as they not only contribute to instruction alignment in the second stage of training but also serve as essential benchmarks for evaluating model performance.

In addition to training datasets, there is a distinct category of datasets that are exclusively designed for evaluating specific capabilities of MLLMs. Examples include TableVQA-Bench~\cite{Kim2024ARXIV_TableVQA_Bench_A}, ChartBench~\cite{Xu2024ARXIV_ChartBench_A_Benchmark}, and MMLongBench-Doc~\cite{Ma2024NEURIPS_MMLongBench_Doc_Benchmarking}. These datasets are tailored to assess advanced functionalities such as long-context understanding, cross-modal reasoning, and domain-specific comprehension. By providing targeted evaluation frameworks, they enable researchers to identify strengths and weaknesses in MLLMs, driving further innovation and refinement in the field.


\section{Challenges and Trends}
\label{sec:summary}


As shown in Table \ref{tab:mllm_summary}, we calculated the average scores from four popular and widely used evaluation datasets, which can basically reflect the performance of MLLMs on TIU tasks. The top five models are Qwen2-VL-72B (88.70), InternVL2.5-78B (87.73), InternVL2.5-38B (87.45), InternVL2.5-26B (85.85), and DeepSeek-VL2-27B (85.40). This indicates that the most state-of-the-art (SOTA) MLLMs currently employ OCR-free encoders, which avoids redundant tokens and complex model architectures. Despite the promising and significant progress made by current MLLMs, the field still faces considerable challenges that require further research and innovation:


\noindent \textbf{Computational Efficiency and Model Compression}. The computational demands of current MLLMs remain a critical bottleneck, primarily due to two factors: (1) the necessity of processing high-resolution document images, which imposes substantial computational resource requirements, and (2) the prevalent use of 7-billion-parameter architectures, while delivering state-of-the-art performance, incur high deployment costs and latency. These challenges underscore the importance of developing more efficient MLLM architectures that balance performance with reduced computational overhead. Encouragingly, recent advancements, as illustrated in Table~\ref{tab:mllm_summary}, demonstrate promising trends toward model miniaturization. For instance, Mini-monkey~\cite{Huang2024ARXIV_Mini_Monkey_Alleviating} achieves performance comparable to 7B-parameter models on multiple TIU tasks while utilizing only 2B parameters, highlighting the potential for lightweight yet powerful architectures.

\noindent \textbf{Optimization of Visual Feature Representation}. A persistent challenge in MLLMs is the disproportionate length of image tokens compared to text tokens, which significantly increases computational complexity and degrades inference efficiency. Addressing this issue requires innovative approaches to compress image tokens without sacrificing model performance. Promising directions include the development of efficient visual encoders, adaptive token compression mechanisms, and advanced techniques for cross-modal feature fusion. Crucially, these methods must preserve the semantic richness of document content during compression. As shown in Table~\ref{tab:mllm_summary}, recent architectural innovations, such as mPLUG-DocOwl2‘s ~\cite{Hu2024ARXIV_mPLUG_DocOwl2_High} visual token compression, have made strides in this direction by enabling the processing of larger input images while maintaining benchmark performance.

\noindent \textbf{Long Document Understanding Capability}. While MLLMs excel at single-page document understanding, their performance on multi-page or long-document tasks remains suboptimal. Key challenges include modeling long-range dependencies, maintaining contextual coherence across pages, and efficiently processing extended sequences. The emergence of specialized benchmarks for long-document understanding~\cite{Ma2024NEURIPS_MMLongBench_Doc_Benchmarking}, as highlighted in Table~\ref{tab:datawithbenchmark}, is expected to drive significant progress in this field by providing standardized evaluation frameworks and fostering targeted research efforts.

\noindent \textbf{Multilingual Document Understanding}. Current MLLMs are predominantly optimized for English and a limited set of high-resource languages, resulting in inadequate performance in multilingual and low-resource language scenarios. Addressing this limitation requires the development of comprehensive multilingual datasets that encompass diverse linguistic and cultural contexts. Recent initiatives, such as MT-VQA~\cite{tang2024mtvqa} and CC-OCR~\cite{yang2024cc} (referenced in Table~\ref{tab:datawithbenchmark}), represent important steps forward by introducing TIU tasks specifically designed to evaluate multilingual capabilities. These efforts, coupled with advances in cross-lingual transfer learning, are expected to significantly enhance the inclusivity and applicability of MLLMs in global contexts.

\section{Limitations}

This paper provides a systematic review of multimodal large language models (MLLMs) in the field of Text-rich Image Understanding (TIU). While the research team has conducted comprehensive retrieval and integration of core literature prior to the submission deadline, certain minor studies may still remain uncovered. It should be particularly noted that due to publisher formatting requirements, the exposition of existing technical approaches and benchmark datasets in this work maintains essential conciseness. For complete algorithmic implementation details and experimental parameter configurations, researchers are strongly recommended to consult the original publications.

\bibliography{main}
\end{document}